\documentclass{article}
\usepackage{arxiv}
\usepackage[utf8]{inputenc} 
\usepackage[T1]{fontenc}    
\usepackage{hyperref}       
\usepackage{url}            
\usepackage{booktabs}       
\usepackage{amsfonts}       
\usepackage{nicefrac}       
\usepackage{microtype}      
\usepackage{lipsum}

\usepackage{algorithm}
\usepackage{algorithmic}
\usepackage{graphicx}
\usepackage{textcomp}
\usepackage{lineno,hyperref}
\modulolinenumbers[5]

\usepackage{graphicx}
\usepackage{times}
\usepackage{xcolor}
\usepackage{soul}
\usepackage[small]{caption}

\usepackage{threeparttable}
\usepackage{multirow}
\usepackage{subfigure}
\usepackage{wrapfig}
\usepackage{authblk}

\title{A Black-box Attack on Neural Networks Based on Swarm Evolutionary Algorithm}
\date{\vspace{-5ex}}

\author[*]{Xiaolei Liu}
\author[*]{Yuheng Luo}
\author[*]{Xiaosong Zhang}
\author[*]{Qingxin Zhu}
\affil[*]{University of Electronic and Science Technology of China}

\begin{document}
\maketitle

\begin{abstract}
Neural networks play an increasingly important role in the field of machine learning and are included in many applications in society. Unfortunately, neural networks suffer from  adversarial samples generated to attack them.  However, most of the generation approaches either assume that the attacker has full knowledge of the neural network model or are limited by the type of attacked model. In this paper, we  propose a new approach that generates a black-box attack to neural networks based on the swarm evolutionary algorithm. Benefiting from the improvements in the technology and  theoretical characteristics of evolutionary algorithms, our approach has the advantages of effectiveness, black-box attack, generality, and randomness. Our experimental results show that both the MNIST images and the CIFAR-10 images can be perturbed to successful generate a black-box attack with 100\% probability on average. In addition, the proposed attack, which is successful on distilled neural networks with almost 100\% probability, is resistant to defensive distillation. The experimental results also indicate that the robustness of the artificial intelligence algorithm is related to the complexity of the model and the data set. In addition, we find that the adversarial samples to some extent reproduce the characteristics of the sample data learned by the neural network model.
\end{abstract}

\section{Introduction}
\label{sec:introduction}

In recent years, neural network models have been widely applied in various fields, especially in the field of image recognition, such as image classification~\cite{li2011effective,yao2018low} and  face recognition \cite{zhao2018novel}. However, users of such model are more concerned about the performance of the model and largely ignore the vulnerability and robustness of the model. In fact, most existing models are easily misled by adversarial samples deliberately designed by attackers and enable the attackers to achieve the purpose of bypassing the detection \cite{szegedy2013intriguing}. For example, in an image classification system, by adding the disturbance information to the original image, attackers can achieve the goal of changing image classification results with high probability \cite{moosavi2016universal}. The generated adversarial samples  can even be classified with an arbitrary label according to the purpose of an attacker, making this type of attack a tremendous threat to the image classification system \cite{barreno2010security}. More seriously, printing the generated images of adversarial samples and then photographing them with a camera, the captured images are still misclassified, confirming the presence of adversarial samples in the real world \cite{kurakin2016adversarial}. These vulnerability problems make people raise the question on whether neural networks can be applied to security-critical areas. 

Several papers have studied related security issues \cite{liu2018improving,liu2018tltd,li2007rough}. Unfortunately, in most previous generation approaches of adversarial samples, when $\epsilon$ is fixed, the similarity of the sample is fixed: in the algorithm's calculation, it won' change dynamically. This may cause the image to be disturbed so much that it can be visually distinguishable \cite{moosavi2017analysis}. Moreover, the existing approaches mainly use gradient information to transform the original samples into the required adversarial samples. If the parameters of the model are unknown, the attackers cannot generate effective adversarial samples \cite{goodfellow2014generative,hu2017generating}. Others also proposed some black-box attack approaches \cite{Papernot2017ACM,Kasiviswanathan2017}. However, Papernot \cite{Papernot2017ACM} takes the transferability assumption. If transferability of the model to be attacked is reduced, the effectiveness of the attack will be reduced. LSA \cite{Kasiviswanathan2017} cannot simply modify the required distance metrics, such as L0, L2, Lmax. In most cases, it is only guaranteed that the disturbance is successful at Lmax, but not guaranteed that the disturbance can be kept minimum under other distance functions.

In this paper, we propose a new approach that generates a \textit{black-box attack} to deep neural networks. Our approach is named BANA, denoting A (B)lack-box (A)ttack on (N)eural Networks Based on Swarm Evolutionary (A)lgorithm.  
Compared with the previous approaches \cite{szegedy2013intriguing,goodfellow2014explaining,papernot2016limitations,carlini2017towards}, our approach has the following main advantages:

\textbf{Effectiveness.} The adversarial samples generated by our approach can misclassify the neural networks with 100\% probability both on non-targeted attacks and targeted attacks. The $L_2$ distance between adversarial samples and original images is less than 10 on average, indicating that images can be disturbed with so small changes that are not to be undetectable. If we continue to increase the number of iterations of our proposed algorithm, we expect to achieve even better results.

\textbf{Black-box Attack.} Adversarial samples can be generated without the knowledge of the internal parameters of the target network, such as gradients and structures. 
Existing attacks such as Carlini and Wagner's attacks~\cite{carlini2017towards} usually require such information. 

\textbf{Generality.} Our proposed attack is a general attack to neural networks. For the attack, we can generate effective adversarial samples of DNNs, CNNs, etc.. We have even tested our proposed attack in a wider range of machine learning algorithms and it still misleads the model with 100\% probability.

\textbf{Randomness.} Benefiting from the characteristics of evolutionary algorithms, the adversarial samples generated each time are different for the same input image, so they are able to resist defensive mechanisms such as defensive distillation.

In particular, our proposed attack is based on the swarm evolutionary algorithm~\cite{coello2007evolutionary}. The swarm evolutionary algorithm is a population-based optimization algorithm for solving complex multi-modal optimization problems. It can transform the optimization problems into the individual fitness function and has a mechanism to gradually improve individual fitness. Evolutionary algorithms do not require the use of gradient information for optimization and do not require that the objective  function be differentiable or deterministic. Different from another approach also based on an evolutionary algorithm \cite{su2017one}, our approach focuses on the optimization of results rather than the number of disturbed pixels. Therefore, we have completely different optimization function and iterative processes from the one pixel attack. Without knowing the parameters of the model, our proposed approach uses the original sample as the input to apply to generate an adversarial sample of the specific label. The used information  is only the probability of the various labels produced by the model. 

Our attack also addresses technical challenges when applying the swarm evolutionary algorithm to generate the adversarial samples. The improvements made in our approach include the optimization of calculation results and convergence speed (see more details in Section 3).

The rest of the paper is organized as follows. Section 2 introduces the related work of adversarial samples. Section 3 presents SEAA (Swarm Evolutionary Algorithm For Black-box Attacks to Deep Neural Networks. Section 4 presents and discusses our experimental results.  Section 5 concludes.

\section{Related Work}
\label{sec:relatedwork}

The adversarial samples of deep neural networks have drawn the attention of many researchers in recent years. \cite{szegedy2013intriguing} used a constrained L-BFGS algorithm to generate adversarial samples. L-BFGS requires that the gradient of the model can be solved,  limiting the diversity of the model and the objective function, and making this approach computationally expensive to generate adversarial samples. \cite{goodfellow2014explaining} proposed the fast gradient sign method (FGSM). However, this approach is designed without considering the similarity of the adversarial samples: the similarity of the generated adversarial  samples may be low. The consequence is that the generated  adversarial  samples may be detected by defensive approaches or directly visually distinguished. An adversarial sample attack named the Jacobian-based Saliency Map Attack (JSMA) was proposed by  \cite{papernot2016limitations}. JSMA also requires the gradient of the model to be solved, and the approach is limited to the $L_0$ distance, and cannot be generated using other distance algorithms \cite{carlini2017towards}. These approaches all assume that the attackers have full access to the parameters of the model. \cite{moosavi2016deepfool} proposed a non-targeted attack approach named Deepfool. This approach assumes that the neural network is linear and makes a contribution to the generation of adversarial samples, while actually neural networks may be not linear. Besides, this approach also does not apply to non-neural network model. Some previous research focused on generating adversarial samples to the malware detection models \cite{yang2017malware,demontis2017yes,grosse2016adversarial}. These adversarial samples also successfully disrupted the model's discriminant results, showing that the common models of machine learning are vulnerable to attacks.

Some recent research aimed to defend against the attack of adversarial samples and proposed approaches such as defensive distillatione \cite{hendrycks2016early,papernot2016distillation,feinman2017detecting,metzen2017detecting}. However, experiment results show that these approaches do not perform well in particular situations due to not being able to defend against adversarial samples of high quality \cite{he2017adversarial}.

\section{Methodology}

\subsection{Problem Description}

The generation of adversarial samples can be considered as a constrained optimization problem. We use $L_p$ distance (which is $L_p$ norm) to describe the similarity between the original images and the adversarial images. Let $f$ be the $m$-class classifier that receives $n$-dimensional inputs and gives $m$-dimensional outputs. Different from L-BFGS \cite{szegedy2013intriguing}, FGS \cite{goodfellow2014explaining}, JSMA \cite{papernot2016limitations}, Deepfool \cite{moosavi2016deepfool} and Carlini and Wagner's attack \cite{carlini2017towards}, our approach is a black-box attack without using the gradient information. This optimization problem is formalized as follows:

\begin{equation}
F = D(x,x') + M \times loss(x')
\end{equation}

where for a non-targeted attack (whose purpose is to mislead the classifier to classify the adversarial samples as any of the error categories), $loss(x')$ is defined as

\begin{equation}
loss(x') = max(\  [f(x')]_{r} - max([f(x')]_{i \neq r})\ , 0)
\end{equation}

and for a targeted attack (whose purpose is to mislead the classifier to classify the adversarial samples as a specified category), $loss(x')$ is defined as

\begin{equation}
loss(x') = max(\   max([f(x')]_{i \neq t}) - [f(x')]_{t}\ , 0)
\end{equation}

and $x=(x_1,...,x_n)$ is the original image, $x'=(x'_1,...,x'_n)$ is the adversarial sample to be produced and $D(x,x')$ is the $L_p$ distance. $M$ is a positive number much larger than $D(x, x')$, $r$ is the real label, and $t$ is the target label. The output of $[f(x)]_r$ is the probability that the sample $x$ is recognized as the label $r$ and the output of $[f(x)]_{i \neq r}$ is the probability set that the sample $x$ is separately recognized as other labels. Since $loss(x') \geq 0$, we discuss the case of $loss(x') > 0$ and $loss(x') = 0$ for the targeted attacks, respectively. The non-targeted attacks are the same.

(1) When $loss(x')> 0$, the $[f(x')]_t$ is not the maximum in Equation 3, indicating that the adversarial sample $x'$ is not classified as the targeted label at this time. Since $M$ is much larger than $D (x, x ')$, the objective function in Equation 1 is approximately equal to the latter half. In this case, it is equivalent to optimizing $x'$ to minimize $Q$, i.e., increasing the probability that the classifier identifies the sample $x'$ as being a class $t$.
\begin{equation}
minimize \  loss(x')
\end{equation}

(2) When $x = 0$, the adversarial sample has been classified as the target label at this time. In this case, it is equivalent to optimizing $x'$  to minimize the value of $D (x, x')$, i.e., to improve the similarity between the adversarial sample and the original sample as much as possible.
\begin{equation}
minimize \ D(x,x')
\end{equation}

\begin{figure}[h]
  \centering
  \includegraphics[width=0.68\textwidth]{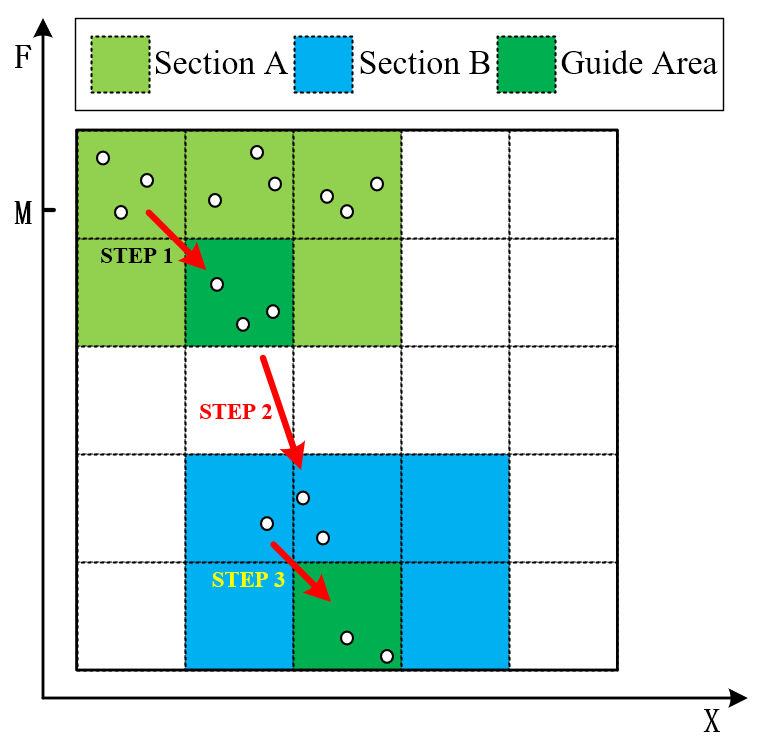}
  \caption{Individuals Distribution Diagram}
  \label{fig-distribution}
\end{figure}

Through the preceding objective function, the population is actually divided into two sections, as shown in  Fig.\ref{fig-distribution}. The whole optimization process can be divided into three steps.

Step 1. At this time the adversarial sample cannot successfully mislead the classifier. Individuals at the top of section A gradually approach the bottom through crossover and mutation operators.

Step 2. The individuals move from Section A  to Section B, indicating that $loss(x')=0$, i.e., the adversarial samples generated at this time can successfully mislead the classifier.

Step 3. Individuals at the top of Section B gradually approach the bottom, indicating the improvement of the similarity between the adversarial image and the original image.

Eventually, the bottom individual of Section B becomes the optimal individual in the population, and the information that it carries is the adversarial sample being sought out.

\subsection{Our BANA Approach}

As the generation of adversarial samples has been considered as an optimization problem formalized as Equation 1, we solve this optimization problem by the swarm evolution algorithm. In this algorithm, fitness value is the result of $F$, population is a collection of $x'$ and many individuals make up the population. By constantly simulating the process of biological evolution, the adaptive individuals which have small fitness value in the population are selected to form the subpopulation, and then the subpopulation is repeated for similar evolutionary processes until the optimal solution to the problem is found or the algorithm reaches the maximum number of iterations. After the iterations, the optimal individual obtained is the adversarial sample $x'$. As a widely applied swarm evolutionary algorithm, such genetic algorithm is flexible in coding, solving fitness, selection, crossover, and mutation. Therefore, in the algorithm design and simulation experiments, we use the following improved genetic algorithm as an example to demonstrate the effectiveness of our BANA approach. The advantages of this approach are not limited to the genetic algorithm. We leave as the future work the investigation of the effects of different types of swarm evolutionary algorithms on our approach.

\subsubsection{Algorithm Workflow}

\begin{figure}[h]
  \centering
  \includegraphics[width=0.95\textwidth]{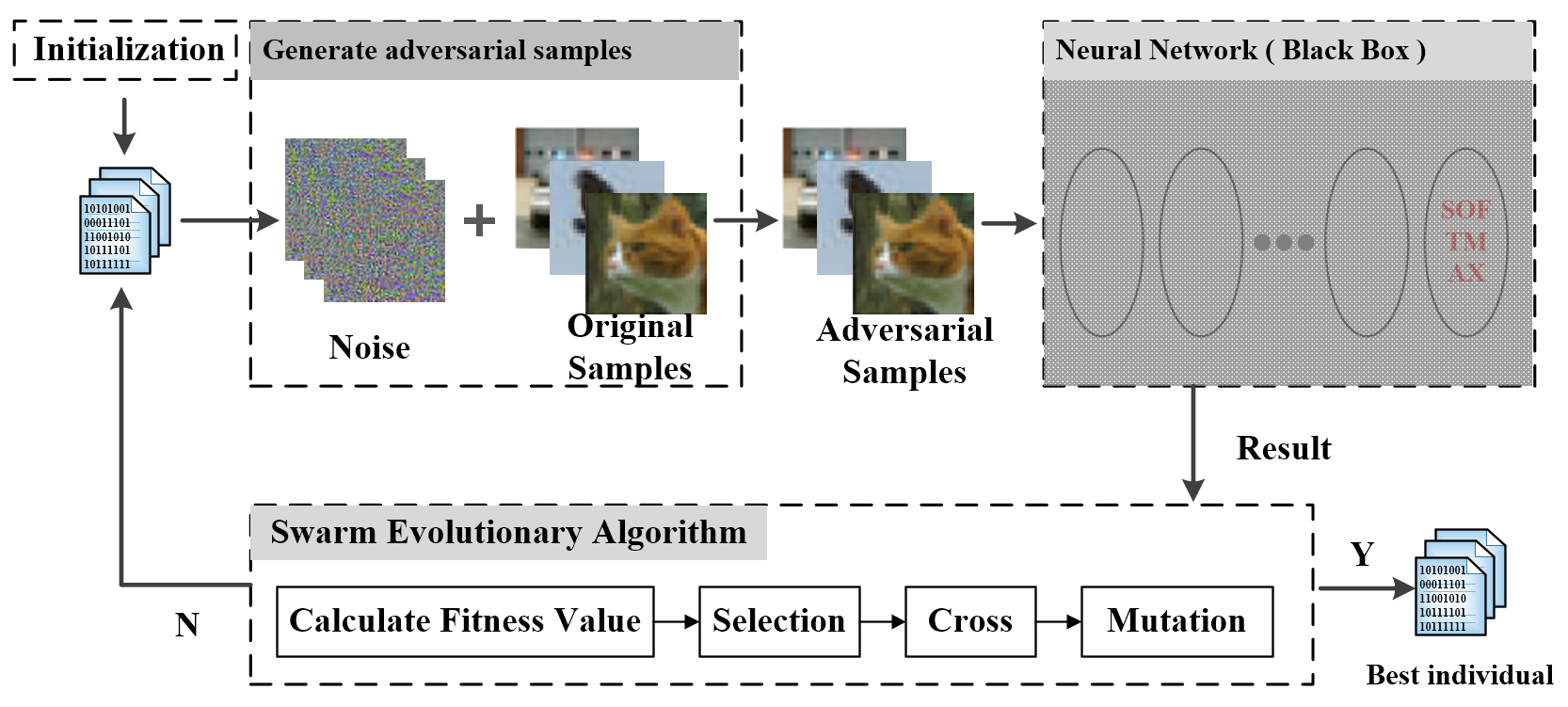}
  \caption{Algorithm Workflow of Our BANA Approach}
  \label{fig-flow-chart}
\end{figure}

The whole algorithm workflow is shown in Fig.\ref{fig-flow-chart}. Classifiers can be logistic regression, deep neural networks, and other classification models. We do not need to know the model parameters and just set the input and output interfaces. Each individual is transformed into an adversarial sample and then sent to the classifier to get the classification result. After that, the individual fitness value is obtained through solving the objective function. The individuals in the population are optimized by the genetic algorithm to solve the feasible solution of the objective function (i.e., the adversarial sample of the image).

The workflow of our BANA approach is as follows:

Step 1. Population Initialization. One gene corresponds to one pixel, and for the grayscale images of (28, 28), there are a total of 784 genes, and there are $32 \times 32 \times 3 = 3072$ genes for the color image of (32, 32).

Step 2. Calculate the Fitness Value. Calculate the value of the fitness function according to the approach described in Section 3.1 and take this value as the fitness of the individual. Since this problem is a minimization problem, the smaller the value, the better the individual's fitness. After that, the best individual with the minimum fitness value in the current population is saved as the optimal solution.

Step 3. Select Operation. According to the fitness of individuals in the population, through the tournament algorithm, individuals with higher fitness are selected from the current population.

Step 4. Cross Operation. Common crossover operators include single-point crossover, multi-point crossover, and uniform crossover. Our algorithm uses uniform crossover. That is, for two random individuals, each gene crosses each independently according to the probability $p$. Due to the large number of genes that each individual carries, uniform crossover allows for a greater probability of generating new combinations of genes and is expected to combine more beneficial genes to improve the searching ability of genetic algorithms.

Step 5. Mutation Operation. In order to speed up the search ability of genetic algorithms, combining with the characteristics of the problem to be solved, the operator adopts a self-defined Gaussian mutation algorithm. In the process of mutation, Gaussian noise $gauss(m, s)$ is randomly added to the individual (shown in Equation 6 below), where $m$ is the mean of Gaussian noise and $s$ is the standard deviation of Gaussian noise: 
\begin{equation}
x_{mutation} = x_{origin} \pm gauss(m,s)
\end{equation}
The reason for adopting this mutation operation is that the resulting adversarial sample inevitably has a high degree of similarity with the input sample, and a feasible solution to the problem to be solved must also be in the vicinity. This technique can effectively reduce the number of iterations required to solve the problem.

Step 6. Terminate the Judgment. The algorithm terminates if the exit condition is satisfied, and otherwise returns to Step 2.

\subsubsection{Improvements}

There are two major technical improvements made in our approach.

\textbf{Improvements of result.} In order to improve the optimization effect of BANA, we adopt a new initialization technique. Considering the problem to be solved requires the highest possible degree of similarity, this technique does not use random numbers while using the numerical values related to the original pixel values. Let $x$ be the original image, and $x'$ be the initialized adversarial image. Then $x' = x + \epsilon$, where $\epsilon$ is a very small value.

\textbf{Improvements of speed.} In order to speed up the convergence of BANA, on  one hand, we constrain the variation step of each iteration in the mutation stage. On the other hand, we try to keep the point that has the pixel value of 0, because it is more likely that such a point is at the background of the picture. These improvements help the algorithm converge faster to the optimal solution.

\section{Experiments}

The datasets used in this paper include MNIST \cite{lecun1998gradient}, CIFAR-10 \cite{krizhevsky2009learning} and ImageNet \cite{Russakovsky2015}. 80\% of the data are used as a training set and the remaining 20\% as a test set. In order to assess the effectiveness of the adversarial samples, we attack a number of different classifier models. The used classifiers  include logistic regression (LR), fully connected deep neural network (DNN), and convolutional neural network (CNN). We evaluate our BANA approach by generating adversarial samples from the MNIST and CIFAR10 test sets.

The parameters used by BANA are shown in Table \ref{tab-parameters}. The experimental results show that different parameters affect the convergence rate of BANA. However, with the increase of the iterations, the results would eventually be close. The parameters listed in Table \ref{tab-parameters} are our empirical values.

\renewcommand{\arraystretch}{1.2} 
\begin{table}[t]
\fontsize{7}{6}\selectfont  
\caption{The parameters of BANA}
\begin{center}
\resizebox{\textwidth}{7mm}{
\begin{tabular}{cccccccc}	\toprule
		Database & Population  & Genes Number & Cross Probability & Mutation Probability & Iterations &Gaussian Mean & Gaussian Variance\cr
	\midrule
		MNIST &   100   &   28$\times$28$\times$1=784 & 0.5 & 0.05 &200 & 0 & 30\cr
		CIFAR-10 &  200 & 32$\times$32$\times$3=3072 & 0.5 & 0.05 &200 & 0 & 20\cr
		ImageNet &  300 & 200$\times$200$\times$3=120000 & 0.5 & 0.05 &100 & 0 & 40\cr
	\bottomrule
\end{tabular}}
\end{center}
\label{tab-parameters}
\end{table}

\subsection{Adversarial Sample Generation on MNIST}
\label{exp1}

\begin{figure*}[htb]
  \centering
  \subfigure[The trend of success rate.] {\includegraphics[width=0.31\textwidth]{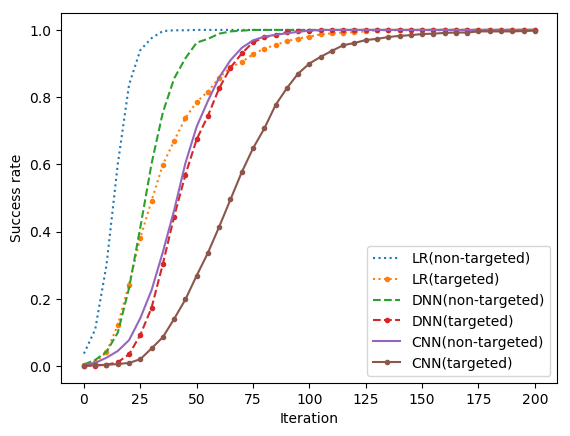} \label{fig-mnist-success}}
  \subfigure[The trend of best fitness value.] {\includegraphics[width=0.31\textwidth]{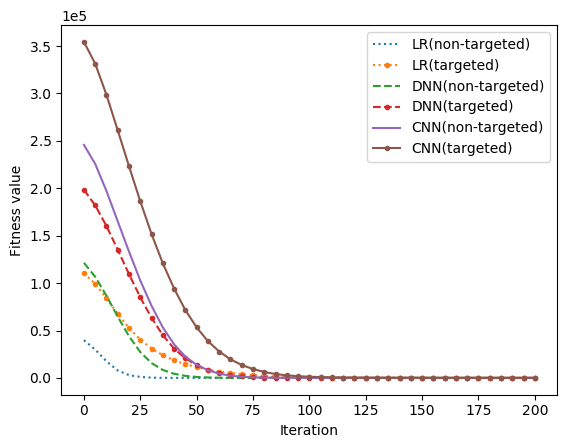} \label{fig-mnist-fitness}}
  \subfigure[The distribution of best fitness.] {\includegraphics[width=0.31\textwidth]{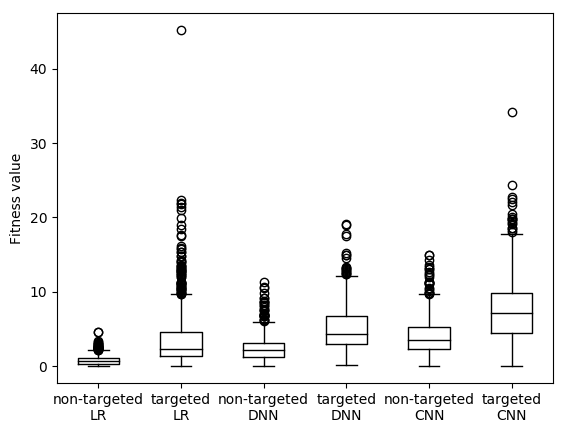} \label{fig-mnist-dist}}
  \caption{The results of targeted attacks and non-targeted attacks for each undistilled model on MNIST.}
  \label{fig-mnist}
\end{figure*}

\begin{figure*}[hbt]
  \centering
  \subfigure[The trend of success rate.] {\includegraphics[width=0.31\textwidth]{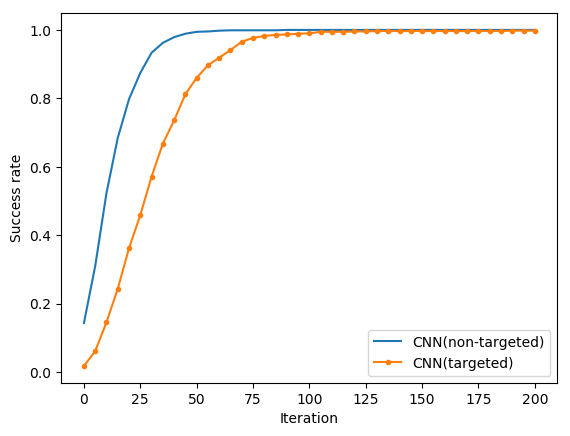}}
  \subfigure[The trend of best fitness value.] {\includegraphics[width=0.31\textwidth]{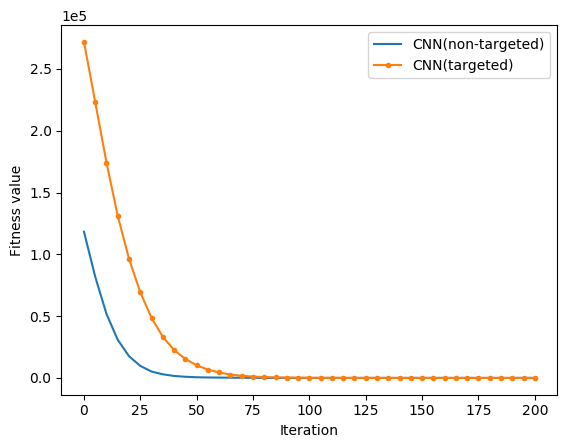}}
  \subfigure[The distribution of best fitness.] {\includegraphics[width=0.31\textwidth]{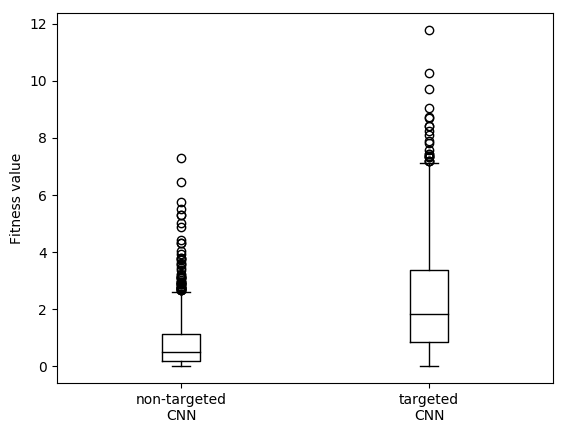}}
  \caption{The results of targeted attacks and non-targeted attacks for each undistilled model on CIFAR-10.}
  \label{fig-cifar}
\end{figure*}

\begin{figure*}[t]
  \centering
  \subfigure[The trend of success rate.] {\includegraphics[width=0.315\textwidth]{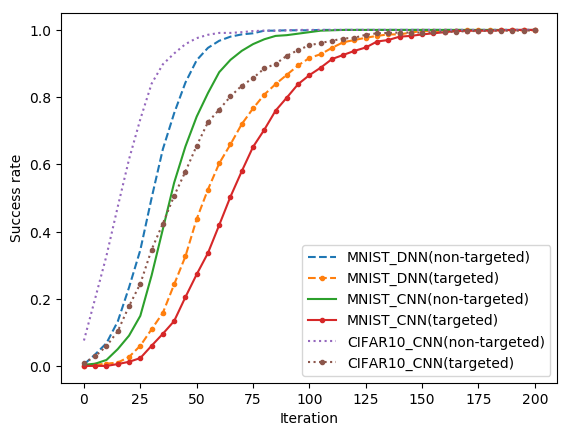}}
  \subfigure[The trend of best fitness value.] {\includegraphics[width=0.315\textwidth]{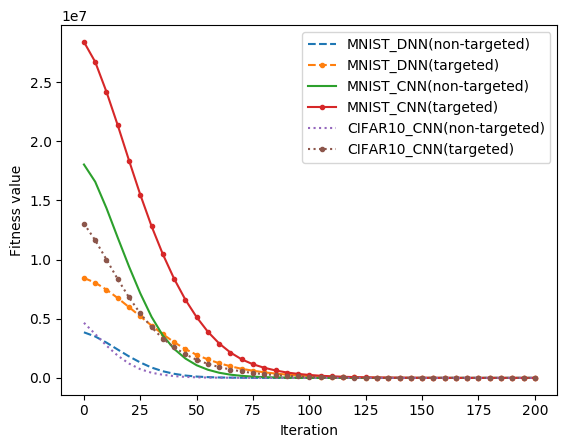}}
  \subfigure[The distribution of best fitness.] {\includegraphics[width=0.31\textwidth]{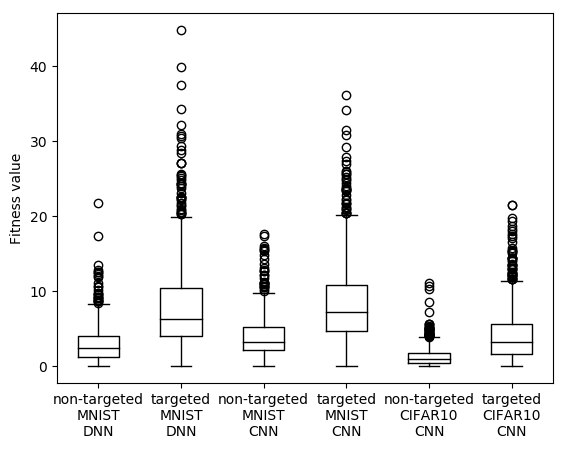}}
  \caption{The results of targeted attacks and non-targeted attacks for each distilled model on MNIST and CIFAR-10. Compare to Fig.~\ref{fig-mnist} and Fig.~\ref{fig-cifar} for undistilled models.}
  \label{fig-distilled}
\end{figure*}

In the first experiment, the used dataset  is MNIST. The used classification models  are LR, DNN, and CNN. We train each of these models separately and then test the accuracy of each model on the test set. Logistic Regression (LR), DNN, and CNN achieve the accuracy of 92.46\%, 98.49\%, and 99.40\% respectively. In the generation of adversarial samples, we set the number of iterations of the genetic algorithm is 200, and the sample with the smallest objective function value generated in each iteration is selected as the optimal sample. For a targeted attack, we select first 100 samples initially correctly classified from the test set to attack. Each of the samples generates adversarial samples from 9 different target labels, resulting in 100 * 9 = 900 corresponding target adversarial samples. For non-targeted attacks, we select the first 900 samples initially correctly classified from the test set to attack. Each sample generates a corresponding adversarial sample, resulting in 900 non-target adversarial samples.

\renewcommand{\arraystretch}{1.2} 
\begin{table}[t]  
  \centering  
  \fontsize{7}{5}\selectfont  
  \begin{threeparttable}  
  \caption{Comparison of our attacks with previous work for a number of MNIST models.
  \\(* The details of distilled model are shown in Section \ref{exp-3}. There is no distilled LR model.
  \\*** This model is attacked by the approach proposed by Carlini and Wagner [2017].)}  
    \begin{tabular}{cccccccccc}  
    \toprule  
    \multicolumn{2}{c}{\multirow{4}{*}{\bf MNIST}}&
    \multicolumn{8}{c}{\bf Models}\cr  
    \cmidrule(lr){3-10} 
    \multicolumn{2}{c}{}&\multicolumn{2}{c}{\bf LR}&\multicolumn{2}{c}{\bf DNN}&\multicolumn{2}{c}{\bf CNN}&\multicolumn{2}{c}{\bf CNN**}\cr
    \cmidrule(lr){3-4} \cmidrule(lr){5-6} \cmidrule(lr){7-8} \cmidrule(lr){9-10} 
    &&{\bf UD(Undistilled)}&{\bf D(Distilled*)}&{\bf UD}&{\bf D*}&{\bf UD}&{\bf D*}&{\bf UD}&{\bf D*}\cr 
    \midrule  
    \multirow{3}{1.2cm}{\bf Non-targeted Attack}& 
{\bf Mean}&0.82&-&2.48&2.91&3.90&3.99&1.76&2.20\cr
&{\bf SD}&0.62&-&1.67&2.34&2.46&2.70&-&-\cr  
&{\bf Prob}&100\%&-&100\%&99.89\%&100\%&100\%&100\%&100\%\cr
    \midrule  
    \multirow{3}{1.2cm}{\bf Targeted Attack}& 
{\bf Mean}&3.65&-&5.04&7.93&7.60&8.33&-&-\cr     
&{\bf SD}&3.80&-&2.88&5.95&4.12&5.34&-&-\cr
&{\bf Prob}&100\%&-&100\%&99.78\%&100\%&100\%&-&-\cr
    \bottomrule  
    \end{tabular}
    \label{tab-mnist}
    \end{threeparttable}  
\end{table}

The results are shown in Table \ref{tab-mnist} and Fig.~\ref{fig-mnist}. For each model, our attacks find adversarial samples with less than 10 in the $L_2$ distance, and succeed with 100\% probability. Compared with the results generated by Carlini and Wagner's attack \cite{carlini2017towards}, our perturbations are slightly larger than their results. However, both of our attacks succeed with 100\% probability and our BANA is a black-box attack. Besides, there is no visual difference between the adversarial samples. Fig.~\ref{fig-mnist-success} and Fig.~\ref{fig-mnist-fitness} show that as the model becomes more complex, the number of iterations required to produce an effective adversarial sample increases. The distribution of the 900 best fitness values after 200 iterations is shown in Fig.~\ref{fig-mnist-dist}. The figure indicates that the more complex the model, the larger the mean and standard deviation. The reason is that simple classification models do not have good decision boundaries. For the same classification model, non-targeted attacks require fewer iterations than targeted attacks, resulting in about $2\times$ lower distortion and stability. Such result indicates that for the attacker the targeted adversarial sample is generated at a higher cost. However, with the increasing of iterations, all the best fitness values tend to be 0. The difficulty caused by the targeted attack can be overcome by increasing the number of iterations. Overall, BANA is able to generate effective adversarial samples for LR, DNN, and CNN on MNIST.

By comparing the trend of success rate and best fitness values for targeted attack and non-targeted attack,  respectively, it can be seen that the robustness of the classification model against adversarial samples is related to the complexity of the model, and the more complex the model, the better the robustness of the corresponding classification model.

\subsection{Adversarial Sample Generation on CIFAR-10 and ImageNet}
\label{exp2}

In the second experiment, the used dataset  is CIFAR-10. Our purpose is to find whether BANA is able to generate effective adversarial samples on CIFAR-10. Considering the conclusion in  Section \ref{exp1}, we choose CNN as the classification model to be attacked. Our CNN achieves an accuracy of 77.82\% on CIFAR-10. After generating the adversarial samples with BANA, we get the results shown in Fig.~\ref{table-cifar}. Our attacks find adversarial samples with less than 2 in the $L_2$ distance and succeed with 100\% probability. We can find the same conclusion as Section \ref{exp1} from Fig.~\ref{fig-cifar} and Table \ref{table-cifar}. 

\begin{figure}[hbt]
  \centering
  \includegraphics[width=0.95\textwidth]{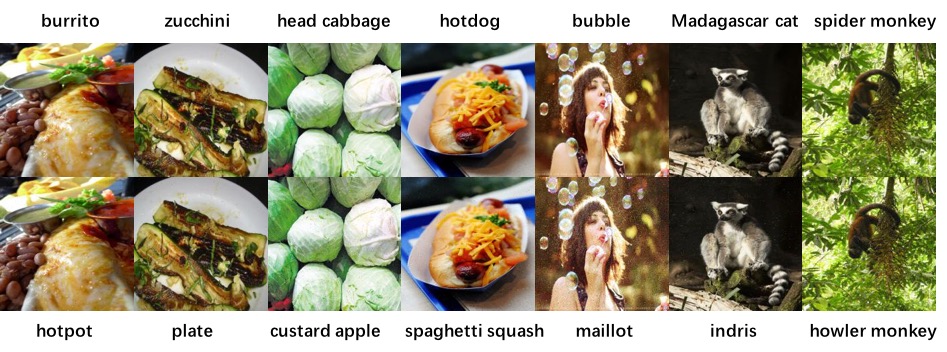}
  \caption{A case study of our BANA on ImageNet. The top row shows the original images and the bottom row shows the perturbed images attacked by our approach.}
  \label{fig-imagenet-example}
\end{figure}

Fig.~\ref{fig-imagenet-example} shows a case study of our BANA on ImageNet. As shown in Fig.~\ref{fig-imagenet-example}, there is no visual difference between the original images and the perturbed images. Fig.~\ref{fig-imagenet-example} shows that our attack is able to generate adversarial samples with small visually invisible perturbations even on complex datasat.
 
\begin{table}[t]  
  \centering  
  \fontsize{7}{5}\selectfont  
  \begin{threeparttable}  
  \caption{Comparison of our attacks with previous work for a number of CIFAR models.\\ (* The details of distilled model are shown in Section \ref{exp-3}.
\\*** This model is attacked by the approach proposed by Carlini and Wagner [2017].)}
\setlength{\tabcolsep}{4mm}{
    \begin{tabular}{cccccccc}  
    \toprule  
    \multicolumn{2}{c}{\multirow{4}{*}{\bf Models}}&
    \multicolumn{6}{c}{\bf CIFAR-10}\cr  
    \cmidrule(lr){3-8} 
    \multicolumn{2}{c}{}&\multicolumn{3}{c}{\bf Non-targeted Attack}&\multicolumn{3}{c}{\bf Targeted Attack}\cr
    \cmidrule(lr){3-5} \cmidrule(lr){6-8} 
    &&{\bf Mean}&{\bf SD}&{\bf Prob}&{\bf Mean}&{\bf SD}&{\bf Prob}\cr 
    \midrule  
    \multirow{2}{*}{\bf CNN}& 
{\bf Undistilled Model}&0.82&0.93&100\%&2.33&1.89&100\%\cr
&{\bf Distilled Model*}&1.26&1.29&100\%&4.18&3.57&99.89\%\cr
\midrule  
    \multirow{2}{*}{\bf CNN**}& 
{\bf Undistilled Model}&0.33&-&100\%&-&-&-\cr
&{\bf Distilled Model*}&0.60&-&100\%&-&-&-\cr
    \bottomrule  
    \end{tabular}}  
    \label{table-cifar}
    \end{threeparttable}  
\end{table}

More importantly, by comparing the experimental results for CNN on MNIST and CIFAR, it can be seen that the average best fitness value and the standard deviation on CIFAR are smaller than them on MNIST, indicating that the adversarial samples generated on CIFAR dataset are more likely to be misleading and more similar to the original data. We find that the robustness of the classification model against adversarial samples is not only related to the complexity of the model but also to the trained data set;  however, not the more complex the data set, the better the robustness of the generated classification model.

\subsection{Defensive Distillation}
\label{exp-3}

We train the distilled DNN and CNN, using $softmax$ at temperature $T = 10$. The experimental results are shown in Tables \ref{tab-mnist}  and~\ref{table-cifar}. The observation is that the average fitness value and standard deviation of undistilled models are smaller than those of distilled model both on targeted attacks and non-targeted attacks. However, the attack success rate of the adversarial samples produced by BABA on the distilled model is still 100\% or close to 100\%. Our attack is able to break defensive distillation. The reason may be related to the randomness of the swarm evolutionary algorithm. Even with the same model and data, BANA produces a different adversarial sample each time, making it effective against defensive distillation.

\subsection{Sample Analysis}

\begin{figure}[hbt]
  \centering
  \includegraphics[width=0.95\textwidth]{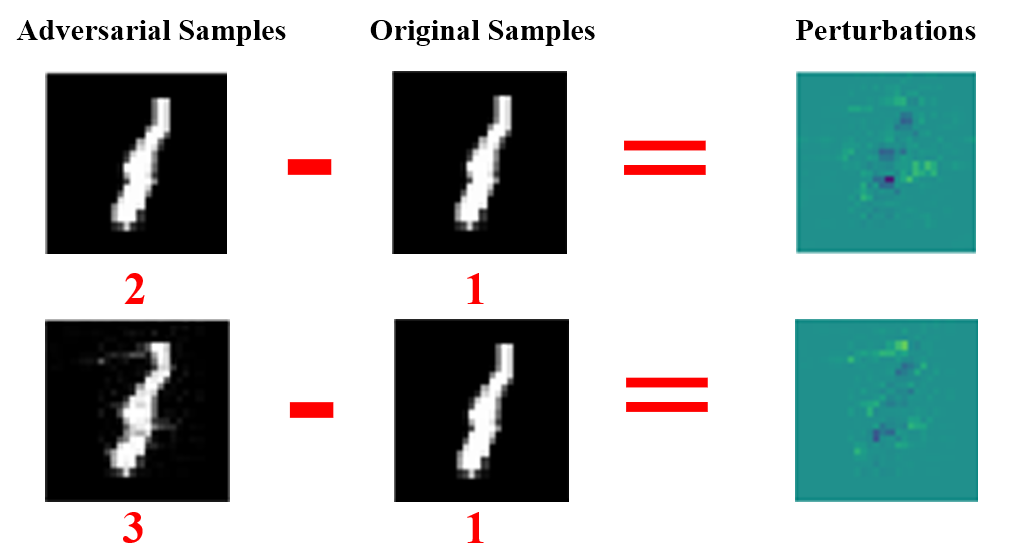}
  \caption{The perturbations of targeted attacks for an MNIST digit.}
  \label{fig-pert}
\end{figure}

The perturbations of targeted attacks for an MNIST digit are shown in Fig.~\ref{fig-pert}. The first column contains the adversarial samples. The second column shows the original samples. The last column shows the perturbations of the targeted samples. The first row is an example of targeted attacks for digit 1 to digit 2. The figure shows that the disturbance in the negative direction is more obvious at the features of digit 1. The disturbance in the positive direction is obvious at the features of the digit 2, and the disturbance area approximates the contour of digit 2. The negative-direction perturbation reduces the probability of the sample being predicted as the real label, and the positive perturbation improves the probability of predicted as a target label. \textbf{Such result indicates  that the adversarial samples to some extent reproduce the characteristics of the sample data learned by the neural networks model.}

\section{Conclusions}

In this paper, we have presented a new approach that generates a black-box attack to neural networks based on the swarm evolutionary algorithm. Our experimental results show that our approach generates high-quality adversarial samples for LR, DNN, and CNN, and our approach is resistant to defensive distillation. Finally, our results indicate that the robustness of the artificial intelligence algorithm is related to the complexity of the model and the complexity of the data set. Our future work includes designing an effective defense approach against our proposed attack.

\bibliographystyle{unsrt}
\bibliography{mybibfile}

\end{document}